\documentclass[letterpaper]{article} % DO NOT CHANGE THIS
\usepackage{aaai23}  % DO NOT CHANGE THIS
\usepackage{times}  % DO NOT CHANGE THIS
\usepackage{helvet}  % DO NOT CHANGE THIS
\usepackage{courier}  % DO NOT CHANGE THIS
\usepackage[hyphens]{url}  % DO NOT CHANGE THIS
\usepackage{graphicx} % DO NOT CHANGE THIS
\usepackage{natbib}  % DO NOT CHANGE THIS AND DO NOT ADD ANY OPTIONS TO IT
\usepackage{caption} % DO NOT CHANGE THIS AND DO NOT ADD ANY OPTIONS TO IT
\usepackage{algorithm}
\usepackage{newfloat}
\usepackage{listings}
\DeclareCaptionStyle{ruled}{labelfont=normalfont,labelsep=colon,strut=off} % DO NOT CHANGE THIS
\floatstyle{ruled}
\newfloat{listing}{tb}{lst}{}
\floatname{listing}{Listing}
\usepackage{algpseudocode}
\usepackage{soul} % strikeout
\usepackage{enumerate}
\usepackage{amsmath}
\usepackage{amssymb}
\usepackage{multirow}
\usepackage{colortbl}
\usepackage{caption}
\usepackage{subcaption}
\usepackage{mathtools}
\usepackage{booktabs}

\title{Anonymization for Skeleton Action Recognition}
\author{
    Saemi Moon\equalcontrib,\textsuperscript{\rm 1}
    Myeonghyeon Kim\equalcontrib,\textsuperscript{\rm 3}
    Zhenyue Qin,\textsuperscript{\rm 4,5}
    Yang Liu,\textsuperscript{\rm 4}
    Dongwoo Kim\textsuperscript{\rm 1,2}\\
}
\affiliations{
    \textsuperscript{\rm 1}CSE, POSTECH $\quad$
    \textsuperscript{\rm 2}GSAI, POSTECH $\quad$
    \textsuperscript{\rm 3}Scatter Lab $\quad$
    \textsuperscript{\rm 4}ANU $\quad$
    \textsuperscript{\rm 5}Tencent\\
    
    \{saemi, dongwookim\}@postech.ac.kr, 
    nessunkim@gmail.com, 
    \{zhenyue.qin, yang.liu3\}@anu.edu.au
}

\begin{document}

\maketitle

\begin{abstract}
Skeleton-based action recognition attracts practitioners and researchers due to the lightweight, compact nature of datasets. Compared with RGB-video-based action recognition, skeleton-based action recognition is a safer way to protect the privacy of subjects while having competitive recognition performance. However, due to improvements in skeleton recognition algorithms as well as motion and depth sensors, more details of motion characteristics can be preserved in the skeleton dataset, leading to potential privacy leakage. 
We first train classifiers to categorize private information from skeleton trajectories to investigate the potential privacy leakage from skeleton datasets. Our preliminary experiments show that the gender classifier achieves 87\% accuracy on average, and the re-identification classifier achieves 80\% accuracy on average with three baseline models: Shift-GCN, MS-G3D, and 2s-AGCN. We propose an anonymization framework based on adversarial learning to protect potential privacy leakage from the skeleton dataset. Experimental results show that an anonymized dataset can reduce the risk of privacy leakage while having marginal effects on action recognition performance even with simple anonymizer architectures. The code used in our experiments is available at \url{https://github.com/ml-postech/Skeleton-anonymization/}
%Also, our proposed framework has better performance than other alternative approaches.
\end{abstract}

\section{Introduction}

%general instruction
Action recognition has been widely studied in many applications such as sports analysis~\cite{tran2018closer}, human-robot interaction~\cite{fanello2013keep}, and intelligent healthcare services~\cite{saggese2019learning}. Due to the success of convolutional neural networks, many recognition approaches are proposed based on a sequence of video frames. Action recognition can further be used for the public good. For example, with surveillance cameras in a public area or a school, we can detect violent actions.

To employ the recognition system appropriately, one must ensure that private information is not abused before and after analysis. Skeleton-based action recognition can be alternative to video-based recognition. Due to the advance in depth and motion sensors, details of motion characteristics can be preserved in the skeleton dataset. Compared with RGB videos, the skeleton dataset seems to expose fewer details on participants. It is often challenging to identify sensitive information such as gender or age from a skeleton to compare with the RGB video to the naked eye. %Skeleton-based action recognition has emerged as a lightweight alternative to video-based action recognition.

%However, convolution over the spatial and temporal dimensions demands excessive computing power. This limits the scalability with the large scale and high-resolution data sets~\cite{qin2021leveraging}. 

% % privacy issue
% \begin{figure}[t!]
% \centering
% \includegraphics[width=\linewidth]{figures/Figure1_v7.pdf}
% \caption{One example of privacy leakage from ETRI-acitivity3D datasets. A gender classifier is employed to identify the gender from a skeleton sequence, exposing the private gender information with high accuracy. After being anonymized, the action classifier is still capable of recognizing the action accurately while the privacy classifier hides the original gender information. Check the detailed experimental setup and results in \autoref{sec:privacy_leakage}.}
% \label{fig:teaser}
% \end{figure}

We raise a question about the privacy-safeness of skeleton datasets. To check potential privacy leakage from skeletons, we conduct experiments on identifying gender or identity with Shift-GCN~\cite{cheng2020skeleton}, MS-G3D~\cite{Liu_2020_CVPR}, and 2s-AGCN~\cite{2019_cvpr_2sagcn}. Based on our analysis, a properly trained classifier can predict private information accurately. Therefore, the skeletons are not safe from the privacy leakage problem. A previous study~\cite{sinhaperson} confirms the possibility of identifying a person from skeletons extracted from Kinect. Also, prior work~\cite{wang2018learning} suggests an end-to-end framework to predict both action and identity recognition tasks.

% previous approaches from different
%For the other types of datasets such as RGB datasets, various approaches have been proposed to remove the private information. \cite{yang2021study} blur a part of an input image to prevent privacy leakage. More advanced techniques such as inpainting methods also proposed to make images unidentifiable. DeepPrivacy \cite{hukkelaas2019deepprivacy} uses pre-trained generative adversarial networks to generate fake faces to replace real ones.
%Although these methods seem promising with RGB datasets, it is unknown whether these approaches are directly applicable to the skeleton dataset. 

\begin{figure}[t!]
     \centering
     \includegraphics[width=\columnwidth]{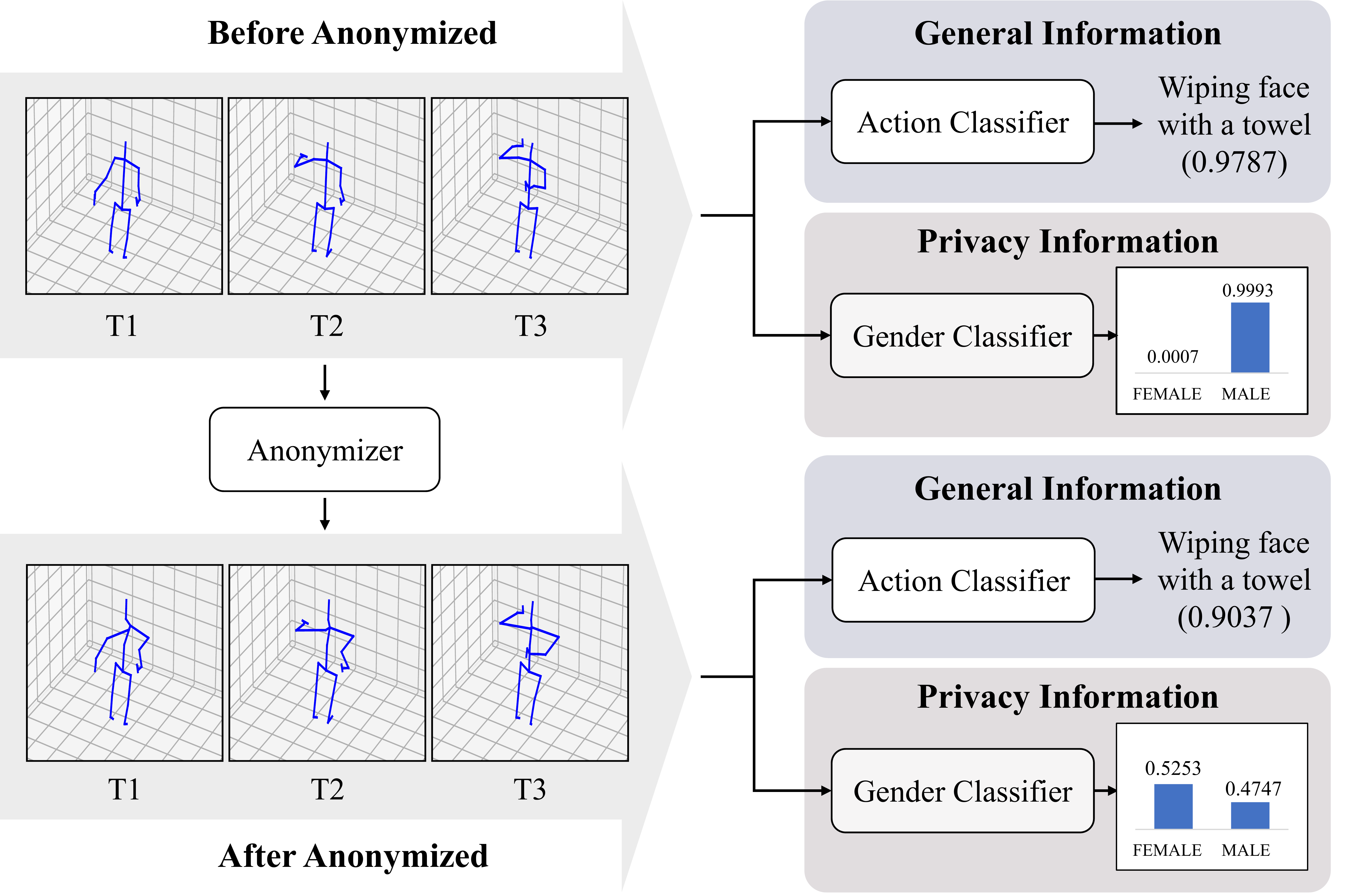}
     \caption{One example of privacy leakage from the ETRI-activity3D dataset. A gender classifier is employed to identify the gender from a skeleton sequence, exposing the private gender information with high accuracy. After being anonymized, the action classifier can still recognize the action accurately while the privacy classifier hides the original gender information.}
\end{figure}

% our approach
This work aims to develop a framework that can anonymize skeleton datasets while preserving critical action features for recognition. To this end, we propose a minimax framework to anonymize the skeletons. With RGB-video datasets, object detection followed by blurring or inpainting with pre-trained generative models is often employed to anonymize datasets~\cite{yang2021study,hukkelaas2019deepprivacy}. However, these methods cannot be directly applied to the skeleton dataset.

The minimax framework consists of an anonymizer network with two sub-networks designed to predict action and private information. The anonymizer removes private information from skeletons, and then the output skeleton is fed into action and privacy classifiers separately. We maximize the accuracy of the action classifier while minimizing the identifiability of private information with the other classifier. In addition, we enforce the anonymized skeleton similar to the original one to make sure they are visually indistinguishable from each other. To solve the minimax problem, we propose an adversarial learning algorithm. Experimental results show that the proposed algorithm results in an effective anonymizer.
\begin{figure*}[hbt!]
\centering
    \includegraphics[width=.8\linewidth]{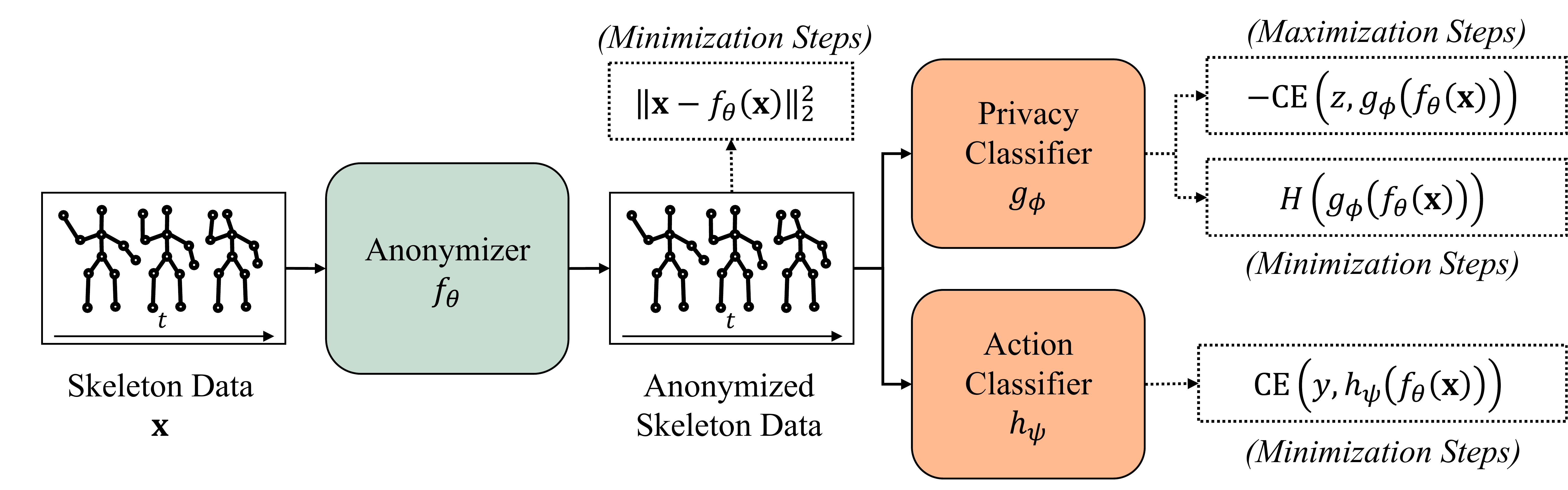}
    \caption{Anonymization framework. The framework consists of three sub-networks: 1) anonymizer $f_\theta$, 2) privacy classifier $g_\phi$, and 3) action classifier $h_\psi$. The dashed box represents the losses used in the minimization and maximization steps with adversarial learning. Note that the privacy classifier uses a separate loss for minimization and maximization in an adversarial learning setup. The parameter of the action classifier $\psi$ is pre-trained and not updated during anonymizer training.}
\label{fig:overall}
\end{figure*}

% contribution
We summarize our contributions as follows:
\begin{itemize}
    \item We empirically show potential privacy leakage from widely-used skeleton datasets such as NTU60~\cite{shahroudy2016ntu} and ETRI-activity3D\cite{jang2020etri}. 
    \item We develop a skeleton anonymization network based on action and sensitive variable classifiers.
    \item We propose a learning algorithm based on the adversarial learning method to anonymize skeletons.
    \item We show that the anonymized skeletons are more robust to privacy leakage while still enjoying high action recognition accuracy.
%    \item We are planning to release an anonymized version of NTU60 dataset\footnote{Full datasets will be available after the peer-review process.}.%, which can be potentially employed for industry usage without disclosing private information leakage.}

\end{itemize}

\section{Skeleton Anonymization}
In this section, we propose a framework for the skeleton anonymization model. %Then, an adversarial learning algorithm is provided to learn the anonymization model.

\paragraph{Anonymization framework.}
Let $\vec{x} \in \mathbb{R}^{T \times D \times 3}$ be 3D coordinates of $D$ joints over $T$ frames, and $y \in \mathcal{Y}$ be an action label for a given skeleton sequence $\vec{x}$, where $\mathcal{Y}$ is a set of actions to be recognized. Let $z \in \mathcal{Z}$ be private information related to the skeleton sequence $\vec{x}$, e.g.,  gender or identity, where $\mathcal{Z}$ is a set of possible private labels.

%A skeleton-based action recognition often aims to identify an action from a trajectory of human joints.
%The goal of an action classifier is to obtain a prediction function $h_\psi:\mathbb{R}^{T \times D \times 3} \rightarrow \mathcal{Y}$ parameterized by $\psi$ from a collection of skeleton trajectories, $D = \{(\vec{x}^{(i)}, y^{(i)}, z^{(i)})\}_{i=1}^{N}$. 

We aim to develop an anonymization network that can effectively remove private information from skeleton datasets while maintaining the recognizability of actions from the anonymized skeletons. To do this, we propose a minimax framework consisting of three different neural network components.
Let $f_\theta: \mathbb{R}^{T \times D \times 3} \rightarrow \mathbb{R}^{T \times D \times 3}$ be an anonymizer network aiming to remove sensitive information from the input skeletons, $h_\psi: \mathbb{R}^{T \times D \times 3} \rightarrow \mathcal{Y}$ be an action classifier, and $g_\phi: \mathbb{R}^{T \times D \times 3} \rightarrow \mathcal{Z}$ be a privacy classifier that predicts sensitive personal information. Our goal is to train an anonymizer $f_\theta$ whose output can maximally confuse the classification performance on the private variables. On the other hand, the output of the anonymizer should keep all relevant information for recognizing action to preserve the performance of the action classifier $h_\psi$. In other words, the output should not be very different from the original skeletons since the anonymized skeletons can be recognizable by the naked eye. To satisfy all requirements, we formalize the anonymization via the following minimax objective:

\begin{align}
    \label{eqn:minimax}
    \begin{split}
        \min_{\theta}\max_{\phi}\mathbb{E}\bigg[\operatorname{CE}\left(y, h_\psi\left(f_\theta(\vec{x})\right)\right) 
        & - \alpha \operatorname{CE}\left(z, g_\phi\left(f_\theta(\vec{x})\right)\right) \\
        & + \beta || \vec{x} - f_\theta(\vec{x}) ||_2^2\bigg]\;,
    \end{split}
\end{align}

where $\operatorname{CE}$ is the cross entropy, and $\alpha$ and $\beta$ are hyperparameters controlling the importance of the privacy classification and the reconstruction error, respectively. The reconstruction error between the original and anonymized skeleton data $|| \vec{x} - f_\theta(\vec{x}) ||_2^2$ ensures the anonymized skeletons are similar to the original ones. To maximize the objective, the private classifier needs to classify the private label $z$ correctly. To minimize the objective, the anonymizer makes the actions easily identifiable by action classifier $h_\psi$ while making the private classifier misclassify the private label $z$.
To simplify the learning process, we use a pre-trained action classifier and fix the parameters of the action classifier during training. The fixed action classifier constrains the anonymized skeleton compatible with the pre-trained model. The anonymized skeletons are also likely to work well with other pre-trained classifiers available.
 
Minimizing the objective w.r.t $\theta$ can make the anonymizer fool the private classifier. %flip the label with a binary classification problem.
However, one may exploit this fact to infer the true label. For example, in a binary classification problem, the true label can be obtained by choosing the opposite of the prediction.
To avoid this issue, we minimize the entropy of classified outputs during the minimization step:
\begin{align}
    \label{eqn:minimize}
    \begin{split}
        \min_{\theta} \mathcal{L}_{\text{adv}} = 
        & \min_{\theta} \mathbb{E}\bigg[\operatorname{CE}\left(y, h_\psi\left(f_\theta(\vec{x})\right)\right) \\
        & - \alpha {H}\left(g_\phi\left(f_\theta(\vec{x})\right)\right) 
         + \beta || \vec{x} - f_\theta(\vec{x}) ||_2^2\bigg]\;,
    \end{split}
\end{align}
where ${H}\left(g_\phi\left(f_\theta(\vec{x})\right)\right)$ is the entropy of the distribution of private labels predicted from the anonymized skeleton. Therefore, the optimal anonymizer yields the most confusing skeletons to the private classifier. In the maximization step, we still maximize the negative cross entropy $- \alpha \operatorname{CE}\left(z, g_\phi\left(f_\theta(\vec{x})\right)\right)$ w.r.t. $\phi$ to train the private classifier. Figure \ref{fig:overall} shows the overall framework for data anonymization.
 
%The minimax objective requires optimizing two sets of parameters $\theta, \phi$ for the anonymizer and privacy classifier, respectively. %We propose two different algorithms to optimize these parameters based on adversarial learning and ensemble learning.

\begin{algorithm}[t!]
\caption{Adversarial Anonymization}\label{alg:adv}
\begin{algorithmic}
\Require Pre-trained classifiers $h_\psi$ and $g_\phi$, $E$: \# of epochs, $m$: minibatch size, $k$: \# of minimization steps
\While{until convergence}
    \For{$t \gets 1$ to $k$}
        \State Sample minibatch of $m$ samples $\{(\vec{x}_i, y_i, z_i)\}_{i=1}^{m}$
        \State Compute $\nabla_\theta \mathcal{L}_{\text{adv}}$ with minibatch
        \Comment{Equation \ref{eqn:minimize}}
        \State Update $\theta \gets \theta - \nabla_\theta \mathcal{L}_{\text{adv}}$
    \EndFor
    \State Sample minibatch of $m$ samples $\{(\vec{x}_i, y_i, z_i)\}_{i=1}^{m}$
    \State Compute $\nabla_\phi \alpha\operatorname{CE}\left(z, g_\phi\left(f_\theta(\vec{x})\right)\right)$ with minibatch
    \State Update $\phi \gets \phi - \nabla_\phi \alpha\operatorname{CE}\left(z, g_\phi\left(f_\theta(\vec{x})\right)\right)$    
\EndWhile
\end{algorithmic}
\end{algorithm}

% \begin{algorithm}[t!]
% \caption{Adversarial Anonymization}\label{alg:adv}
% \begin{algorithmic}
% \Require Pre-trained classifiers $h_\psi$ and $g_\phi$, $E$: \# of epochs, $m$: minibatch size, $k$: \# of minimization steps
% \While{until convergence}
%     \For{$t \gets 1$ to $k$}
%         \State Sample minibatch of $m$ samples $\{(\vec{x}_i, y_i, z_i)\}_{i=1}^{m}$
% %        \If{$t$ in Minimization Steps}
%         \State Compute $\nabla_\theta \mathcal{L}_{\text{adv}}$ with minibatch
%         \Comment{\ref{eqn:minimize}}
%         \State Update $\theta \gets \theta - \nabla_\theta \mathcal{L}_{\text{adv}}$
% %        \Else
% %        \EndIf
%     \EndFor
%     \State Sample minibatch of $m$ samples $\{(\vec{x}_i, y_i, z_i)\}_{i=1}^{m}$
%     \State Compute $\nabla_\phi \alpha\operatorname{CE}\left(z, g_\phi\left(f_\theta(\vec{x})\right)\right)$ with minibatch
%     \State Update $\phi \gets \phi - \nabla_\phi \alpha\operatorname{CE}\left(z, g_\phi\left(f_\theta(\vec{x})\right)\right)$    
% \EndWhile
% \end{algorithmic}
% \end{algorithm}

%\paragraph{Learning algorithm.
Alternating minimization and maximization are often employed to solve a minimax objective as shown in the generative adversarial network~\cite{goodfellow2014generative}. Following previous work, we also use the alternating algorithm to optimize the objective. Algorithm \ref{alg:adv} shows the overall training algorithm. In this work, the adversarial learning algorithm starts with pre-trained classifiers $g_\phi$ and $h_\psi$ to make the learning stable.

%It is known that the minimax objective is difficult to optimize in general. 
%  \sm{Empirically, we obtain a better anonymizer with the pre-trained models.}

% \For{$k \gets 1$ to $E$}
%     \State Sample minibatch of $m$ samples $\{(\vec{x}_i, y_i, z_i)\}_{i=1}^{m}$
%     \State Compute $\nabla_\theta \mathcal{L}_{\text{adv}}$ with minibatch
%     \Comment{\ref{eqn:minimize}}
%     \State Update $\theta \gets \theta - \nabla_\theta \mathcal{L}_{\text{adv}}$
% %    \State Sample minibatch of $m$ samples $\{(\vec{x}_i, y_i, z_i)\}_{i=1}^{m}$    
% %    \State Compute $\nabla_\psi \mathcal{L}_{\text{adv}}$ with minibatch
% %    \Comment{\ref{eqn:minimize}}    
%     % \State Update $\psi \gets \psi - \nabla_\psi \mathcal{L}_{\text{adv}}$
%     \State Sample minibatch of $m$ samples $\{(\vec{x}_i, y_i, z_i)\}_{i=1}^{m}$    
%     \State Compute $\nabla_\phi \alpha\operatorname{CE}\left(z, g_\phi\left(f_\theta(\vec{x})\right)\right)$ with minibatch
%     \State Update $\phi \gets \phi - \nabla_\phi \alpha\operatorname{CE}\left(z, g_\phi\left(f_\theta(\vec{x})\right)\right)$
% \EndFor

\paragraph{Anonymizer networks.}
The anonymizer $f_\theta$ can be any prediction model that modifies skeletons while preserving the original dimension. We employ two simple neural network architectures for the anonymizer: 1) residual networks and 2) U-net architectures.

First, the residual network~\cite{he2016deep} anonymizer adopts a simple residual connection from the input skeletons to the output skeletons. Specifically, the model can be formalized as
\begin{align*}
    f_\theta(\vec{x}) = \operatorname{MLP}_\theta(\vec{x}) + \vec{x},
\end{align*}
where $\operatorname{MLP}_\theta:\mathbb{R}^{D \times 3} \rightarrow \mathbb{R}^{D \times 3}$ is a simple multi-layered perceptron parameterized by $\theta$. The residual connection keeps the position similar to the original skeleton while the $\operatorname{MLP}$ layer models the disposition of joints to anonymize. We use two fully-connected layers to model the disposition. The anonymizer is applied to each frame of a skeleton sequence. Although the anonymizer is applied to each frame independently, the back-propagated signals from action and private classifiers make the entire sequence coherent. By initializing $\theta$ with weights close to zero, we make the anonymizer add a small random noise to the original skeleton in the early stage of learning.

Second, the U-Net~\cite{ronneberger2015u} architecture is adopted to our anonymizer network. The U-Net consists of two paths: the contracting path, and the expanding path. In the contracting path, it repeats downsampling and maxpool an input skeleton data to encode it to the feature map. In the expanding path, U-Net repeats upsampling and concatenating feature maps via skip connection. Especially, skip connections concatenates the features from the contracting path to the corresponding level in the expanding path. It makes the output skeleton position similar to the original skeleton. The detailed architecture is provided in Appendix.
%This (the detailed architecture) can go to appendix: \footnote{We adopt the U-Net model from this repository: https://github.com/milesial/Pytorch-UNet}
% \begin{align*}
%     f_\theta(\vec{x}) = \operatorname{Dec}(\operatorname{Enc}(\vec{x})),
% \end{align*}
% where $\operatorname{Dec}:\mathbb{R}^{D \times 3} \rightarrow \mathbb{R}^H$ and $\operatorname{Enc}: \mathbb{R}^H \rightarrow \mathbb{R}^{D \times 3}$ are the decoder and encoder parameterized by $\theta$, where $H$ is the size of the latent vector. In this work, we use two fully-connected layers for both encoder and decoder. To train the auto-encoder anonymizer, we pre-train the model by reconstructing the skeleton with a training dataset. The encoder layers are then fixed, and only the decoder layers are trained by \ref{alg:adv}.

 %For example, one can employ well known auto-encoder architectures such as the variational auto-encoder \cite{kingma2014autoencoding} as an anonymizer.

\section{Experiments}
In this section, we demonstrate the performance of the proposed framework for anonymizing skeleton datasets. We use two publicly available datasets and anonymize two different types of private information: gender and identity.

\begin{figure*}[t!]
\centering
\begin{subfigure}[t]{0.4\textwidth}
\includegraphics[width=\linewidth]{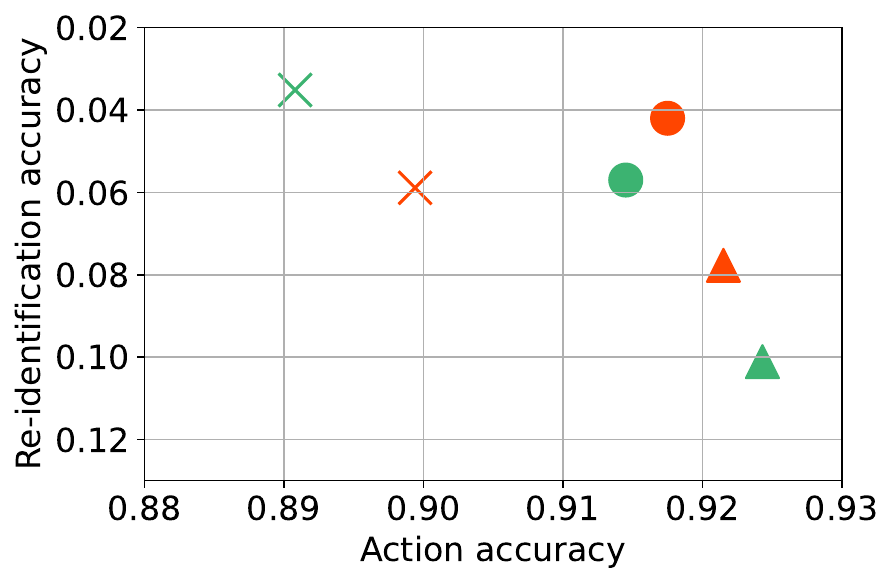}
\caption{Re-identification task}
\end{subfigure}
\begin{subfigure}[t]{0.4\textwidth}
\includegraphics[width=\linewidth]{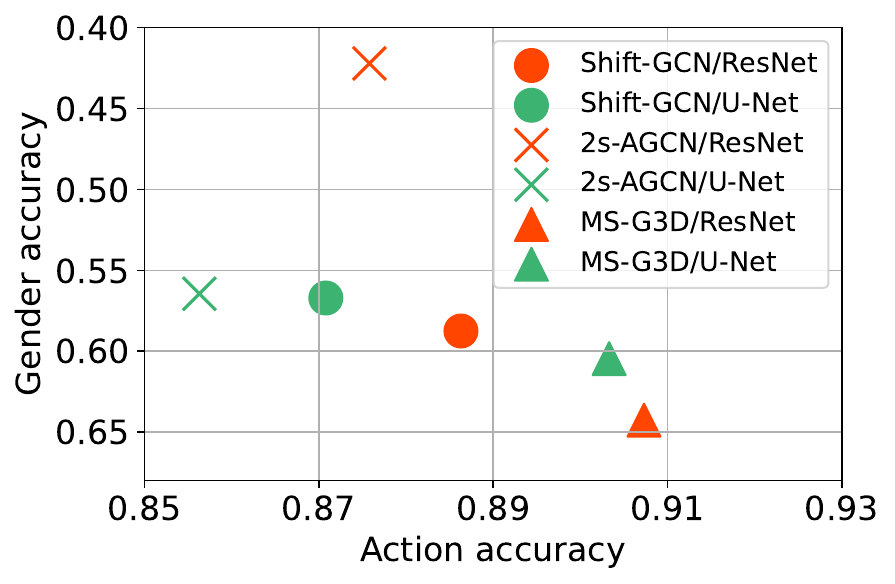}
\caption{Gender classification}
\end{subfigure}
\caption{Action and privacy accuracy of three baseline models with two different anonymizers after anonymization. $y$-axis is reversed. Note that, before anonymization, the average top-1 re-identification accuracy is 80\%, and the average gender classification accuracy is 87\%.}
\label{fig:final_results}
\end{figure*}

\subsection{Datasets}
\label{sec:datasets}

We use two datasets: ETRI-activity3D~\cite{jang2020etri} and NTU RGB+D 60 (NTU60)~\cite{shahroudy2016ntu}. For the ETRI-activity3D dataset, we anonymize the gender information from the skeletons. For the NTU60 dataset, we anonymize the identity of the skeletons. The detailed experimental setups for these datasets are as follows.

\paragraph{ETRI-activity3D.}
ETRI-activity3D is an action recognition dataset originally published for recognizing the daily activities of the elderly and youths. It contains 112,620 skeleton sequence samples observed from 100 people, half of whom were between the ages of 64 and 88 and the rest were in their 20s. The elderly consist of 33 females and 17 males, and the young adults consist of 25 females and 25 males. The samples are categorized into 55 classes based on the activity type. Each action is captured from 8 different Kinect v2 sensors to provide multiple views. Each sequence consists of 3D locations of 25 joints of the human body.

With the ETRI-activity3D dataset, we anonymize the gender information from the skeletons. We drop samples from 5 classes for the following experiments, e.g., handshaking, containing two people, so only one person appears in the remaining samples. After removing malformed and two-person samples, we split the remaining samples into 68,788 and 34,025 training and validation, respectively. We split the dataset according to the subject ID. In other words, the subjects in the validation set do not appear in the training set. Through this split, we measure the generalizability of the gender classifier to the unknown subjects.

\paragraph{NTU60.}
NTU60 is an action recognition dataset that contains 60 action classes and 56,880 skeleton sequences taken from 40 subjects. The format of skeleton data is the same as ETRI-activity3D, which includes the 3D positions of 25 human body joints. After removing malformed samples, we split the remaining samples into 37,646 and 18,932 as training and validation sets, respectively. Following the original work \cite{shahroudy2016ntu}, we split it according to the camera ID so that both sets contain identical subjects with different views.

\begin{figure*}[t!]
\centering
\begin{subfigure}[t]{0.4\textwidth}
\includegraphics[width=\linewidth]{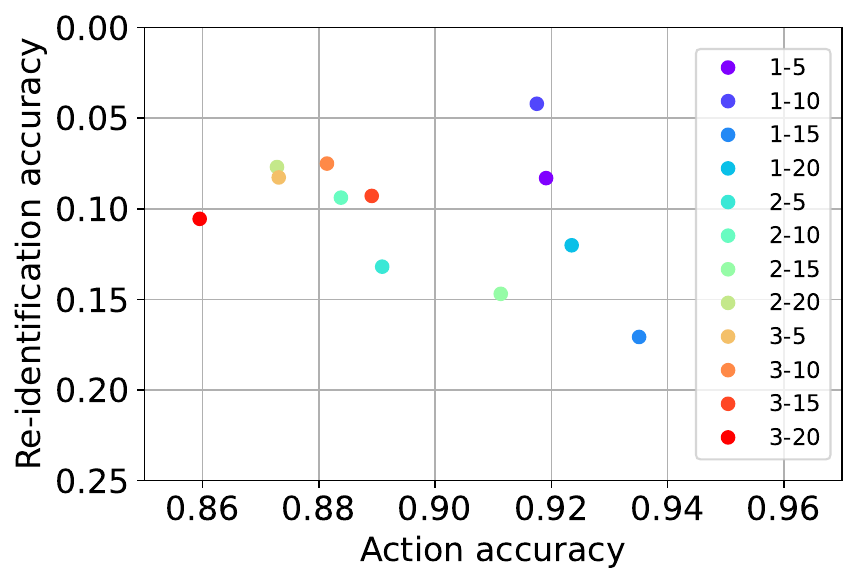}
\caption{\label{fig:ntu_grid_resnet}NTU60 - ResNet}
\end{subfigure}
\begin{subfigure}[t]{0.4\textwidth}
\includegraphics[width=\linewidth]{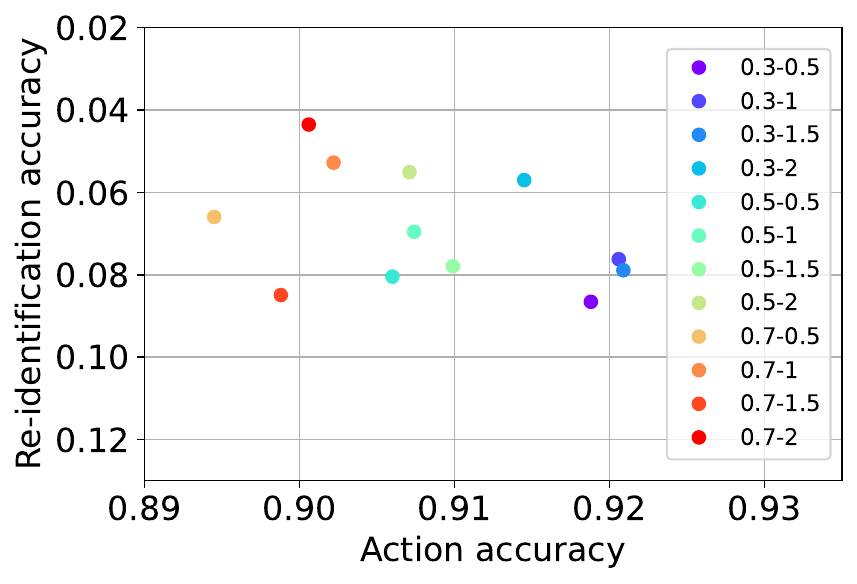}
\caption{\label{fig:ntu_grid_unet}NTU60 - U-Net}
\end{subfigure}

\begin{subfigure}[t]{0.4\textwidth}
\includegraphics[width=\linewidth]{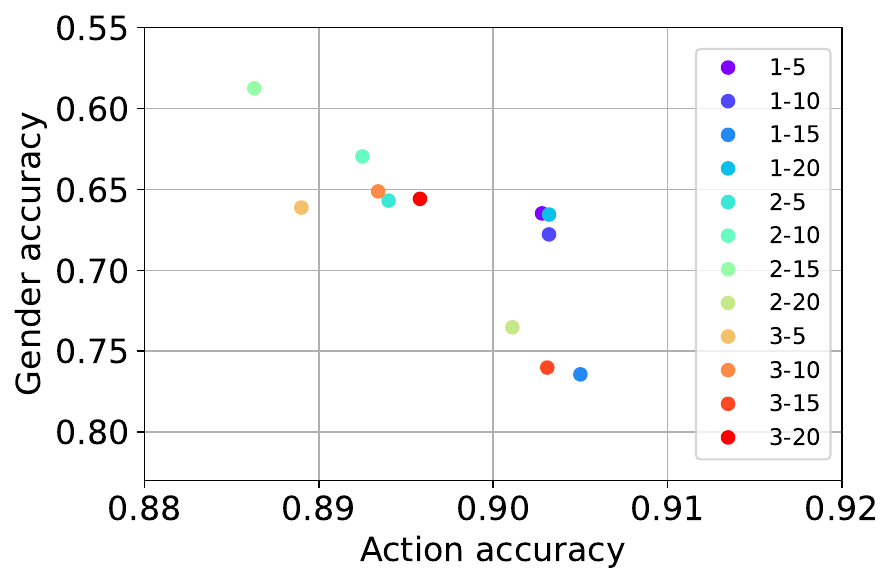}
\caption{\label{fig:etri_grid_resnet}ETRI-activity3D - ResNet}
\end{subfigure}
\begin{subfigure}[t]{0.4\textwidth}
\includegraphics[width=\linewidth]{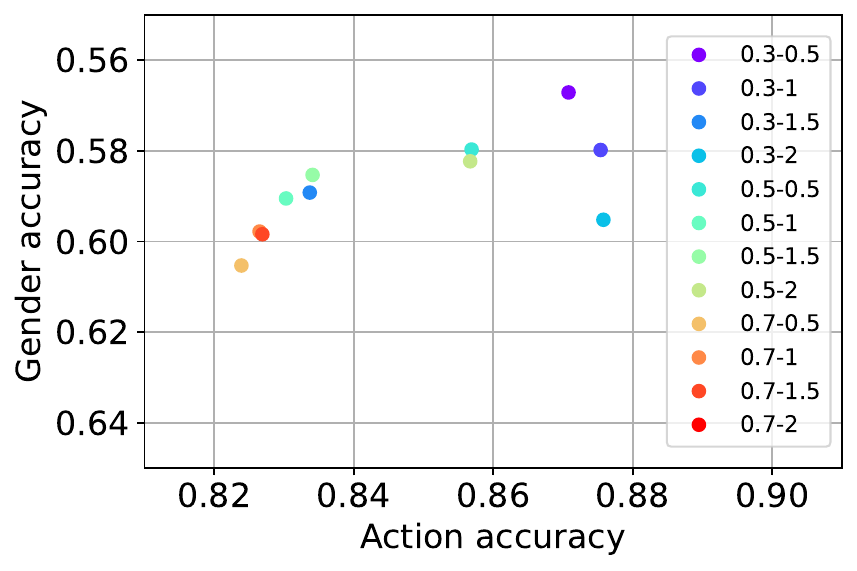}
\caption{\label{fig:etri_grid_unet}ETRI-activity3D - U-Net}
\end{subfigure}
\caption{The trade-off between action accuracy and privacy accuracy based on a different configuration of hyperparameter $\alpha$ and $\beta$ on NTU60 and ETRI-activity3D with two anonymizer networks (legend: $\alpha$-$\beta$). Note that the y-axis is reversed.}
\label{fig:ntu_grid_result}
\end{figure*}

\begin{figure*}[t!]
\centering
\begin{subfigure}[t]{0.4\textwidth}
\includegraphics[width=\linewidth]{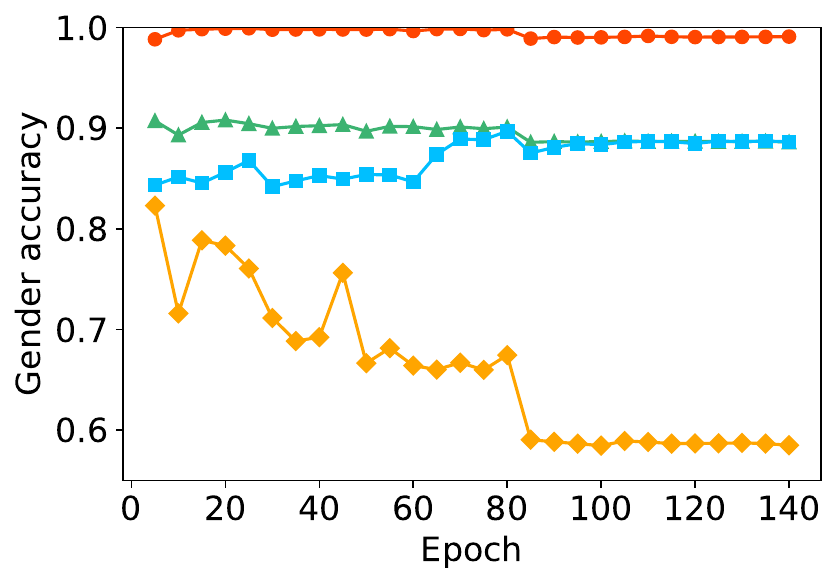}
\caption{\label{fig:train_vs_val_gender}Gender classification}
\end{subfigure}
\begin{subfigure}[t]{0.4\textwidth}
\includegraphics[width=\linewidth]{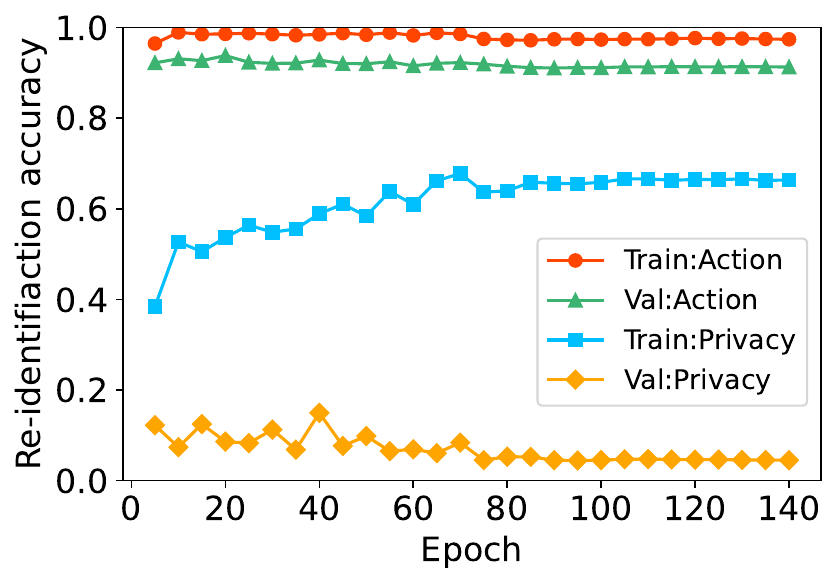}
\caption{\label{fig:train_vs_val}Re-identification task}
\end{subfigure}

\caption{Accuracy and reconstruction error over epochs with the residual anonymizer. `Train:Action' and `Val:Action' indicate the training and validation accuracy of the action classification, and `Train:Privacy' and `Val:Privacy' indicate the training and validation accuracy of the privacy classification.}
\label{fig:adv_trace}
\end{figure*}

\subsection{Privacy Leakage}
\label{sec:privacy_leakage}

To verify privacy leakage from each dataset, we first check the performance of the gender classification and re-identification task. To train gender classifier and re-identification task, three popular baseline models, Shift-GCN~\cite{cheng2020skeleton}, MS-G3D~\cite{Liu_2020_CVPR}, and 2s-AGCN~\cite{2019_cvpr_2sagcn}, are adopted. For the gender classifier with MS-G3D, we use MS-G3D without a G3D module. This makes training faster without losing too much accuracy. We train multiple times for the gender classifier and re-identification task. Each model is trained with a different random initialization.

After training, the gender classifier achieves 87\% accuracy on average. The re-identification task achieves 80\% and 97\% for top-1 and top-5 accuracy, respectively. The detailed results are available in Appendix. As the results suggest, the privacy information can be easily predicted by a classification model trained with private labels. Note that the test splits do not contain the subject used for gender classification. This reveals the generalizability of gender classification to unseen subjects. Also, for the re-identification task, the train split and test split have different camera IDs, so the same person appears in both sets with different views. This result indicates that the joint trajectory contains personal traits that can be easily exploited to identify a person.

\subsection{Anonymization Results}
\label{sec:anonymization_results}

Our preliminary study indicates that gender and identification can leak enough from training. Based on the results obtained in the previous experiments, we evaluate the performance of anonymization with an adversarial learning algorithm. As mentioned earlier, experiments are conducted with two anonymizer networks (ResNet~\cite{he2016deep}, U-Net~\cite{ronneberger2015u}), two datasets (NTU60~\cite{shahroudy2016ntu}, ETRI-activity3D~\cite{jang2020etri}), and three baseline models (Shift-GCN~\cite{cheng2020skeleton}, MS-G3D~\cite{Liu_2020_CVPR}, and 2s-AGCN~\cite{2019_cvpr_2sagcn}). %Three baseline models are applied for action classification and privacy classification.
For each task, we use two pre-trained classifiers for action and privacy, respectively. One classifier is used to initialize the adversarial algorithm, and the other is used to measure the accuracy after anonymization.

Figure \ref{fig:final_results} shows the results of anonymization with re-identification task and gender classification. In general, we observe that we can dramatically decrease privacy accuracy while minimally sacrificing action recognition accuracy. We also observe more leakage of private information when the action accuracy is relatively higher. 
Note that we use a balanced test set for identity anonymization. Since NTU60 has 40 subjects, one can achieve 2.5\% accuracy with random classification.
For the gender classification, one can achieve 50\% accuracy with a random classifier on the test set.
%\dw{The gender classification test set consists of 60.46\% of males. Hence, one can achieve 50\% accuracy with random classifier on the test set.}
%This means that our anonymizer framework works properly to anonymize privacy information. \dw{need to say something about the difference between models.}

One would expect the trade-off between action accuracy and privacy accuracy based on the choice of hyperparameters $\alpha$ and $\beta$. The choice of the best anonymization model may vary depending on the application. In this work, we report the performance of the best model based on \textit{action accuracy} $\times$ \textit{(1 - re-identification accuracy)} from the results of various configurations. According to our metric, ResNet ($\alpha$:1, $\beta$:10) and  U-Net ($\alpha$:0.3, $\beta$:2) models are chosen as representative model. We provide additional results with different configurations in the next section. Note that one may use a different metric to select a model given different application scenarios. 
We use the best configuration obtained from the baseline model, Shift-GCN, to train the other baseline models.
The detailed configuration of hyperparameters used to plot Figure \ref{fig:final_results} is available in Appendix.

\paragraph{Trade-off Analysis.}

We vary the value of $\alpha$ and $\beta$ to check the trade-off between action accuracy and privacy leakage based on different configurations of hyperparameters. We use Shift-GCN as a baseline model for analysis. %with fixed learning rate of 0.01.
Figure \ref{fig:ntu_grid_resnet} and Figure \ref{fig:ntu_grid_unet} show the result with various hyperparameter configurations on the identity anonymization task. Figure \ref{fig:etri_grid_resnet} and Figure \ref{fig:etri_grid_unet} show the result of the gender anonymization task. We can observe that given a fixed $\alpha$, increasing $\beta$ increases the chance of privacy leakage as well as the action accuracy showing the presence of the trade-off between the action accuracy and privacy leakage.

%it is hard to choose the best model, because action accuracy and privacy accuracy are considered at the same time. Therefore, we set new metric that can help choose the best model by calculating action accuracy and (1 - privacy accuracy). To choose better $\alpha$ and $\beta$, grid search is conducted with three different $\alpha$ and four different $\beta$ in Shift-GCN. After finding them, similar $\alpha$ and $\beta$ are applied to other baselines like 2s-AGCN and MS-G3D.

%As shown in \ref{tab:anonymize_identity}, the residual network anonymizer significantly drops the re-identification accuracy from 83\% to 8\% while keeping the action classification accuracy relatively high.

% After anonymization, the holdout action classifier achieves an accuracy of 86.02\%, and the re-identity classifier achieves an accuracy of 6.03\%. 

We additionally plot how the validation accuracy changes over training procedure with adversarial learning in Figure \ref{fig:adv_trace}. For both datasets, the action accuracy remains high over the epochs on both training and validation sets. However, there is a gap between the training and validation accuracy on privacy, which shows the overfitting in the classifier. Specifically, for gender classification, gender accuracy starts with similar values. Then gender accuracy increases on the train set and decreases on the validation set. The re-identification accuracy drops first on both training and validation sets for the re-identification task. Note that the re-identification task achieves 80\% accuracy. The validation accuracy at the first epoch indicates that the re-identification task is more sensitive to the additional noise introduced by random weights of the residual network than the gender classification.

\paragraph{Reconstruction Error Analysis.}

\begin{figure}[h]
\centering
\begin{subfigure}[b!]{0.43\textwidth}
    \includegraphics[width=\linewidth]{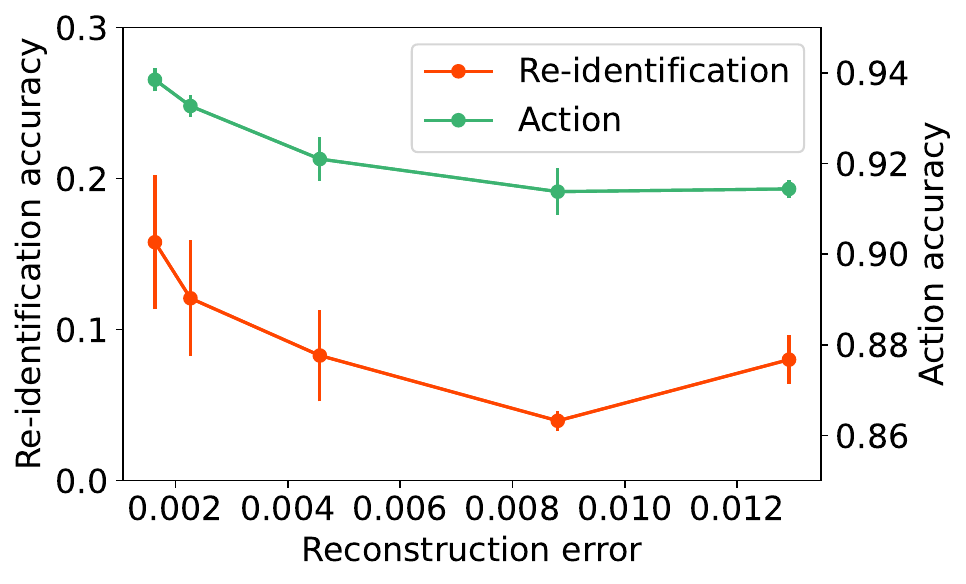}
    \caption{The trade-off between reconstruction error and accuracy with the residual anonymizer on NTU60.}
\label{fig:recon_analysis}
\end{subfigure}
\end{figure}

A reconstruction error directly shows the difference between the original and anonymized skeletons. Although we cannot directly set the level of reconstruction error, we vary the parameters to obtain different levels of reconstruction error and corresponding prediction accuracy. Please check the Appendix for the detailed hyperparameter settings. As shown in Figure \ref{fig:recon_analysis}, 
there is a trade-off between reconstruction error and re-identification accuracy. As we increase the reconstruction error, we can reduce the re-identification accuracy. However, high reconstruction error yields low action accuracy as well. 
%\begin{figure}[t]

%\end{figure}
%if the reconstruction error is relatively high, both action and gender classifiers perform poorly. When the reconstruction error $\approx 0.05$, the skeletons can preserve action information while gender information is anonymized.
%With a small reconstruction error ($\approx 0.01)$, we cannot remove the private information from the skeletons.
%The gender classification accuracy is relatively unstable compared with the action classification accuracy. We conjecture due to subtle cues of gender from skeletons, the anonymization depends on initialization a lot.

% \begin{table}
% \begin{center}
% \begin{tabular}{ l | r r r} 
%  \toprule
%  Model & Recon. & Action & ReID \\
%  \midrule
%  Not-anonymized & 0.0000 & 0.9496 & 0.8284 \\ 
% % Random noise & 0.4740 & 0.3419 & 0.1006 \\ 
%  \midrule
%  Auto-encoder & 0.1279 & 0.6153 & 0.0996 \\
%  ResNet & 0.1184 & 0.8125 & 0.0789 \\ 
%  \bottomrule
% \end{tabular}
% \caption{Reconstruction loss, action accuracy, and re-identification accuracy before and after anonymization. We anonymize the skeletons with adversarial learning. Ens. denotes the ensemble learning method, and Adv. denotes the adversarial learning method.}
% \label{tab:anonymize_identity}
% \end{center}
% \end{table}

\paragraph{Comparison with Alternative Approaches.}
Since we propose privacy leakage for the first time, no anonymization method removes privacy information while remaining action accuracy high. Therefore, we consider two alternative approaches to anonymize privacy information by modifying skeleton data potentially. 
(1) Random noise: As a baseline, we randomly inject white noise drawn from the zero mean normal distributions with varying variances to the original skeleton. 
(2) Adversarial attack method: Adversarial attack is a technique that makes a model fool by perturbing input data. There are several adversarial attack research on skeleton action recognition \citep{liu2020adversarial, wang2021understanding, diao2021basar, tanaka2022adversarial, zheng2020towards}. We use \citet{wang2021understanding} method to attack privacy information. Note that we select Shift-GCN as a baseline model in this experiment.

Table \ref{tbl:compare} shows the results of comparing our method to other approaches.
The random noise cannot preserve action information while reducing privacy leakage. 
The results with an adversarial attack show that attacked skeleton data succeeds in removing privacy information from the target model, i.e., the identification accuracy of the attacked model is zero. However, identification accuracy remains relatively high for the other pre-trained model, which has not been attacked. The adversarial attack-based anonymization is model-specific and difficult to generalize to the unseen models, whereas anonymized skeleton data with our proposed framework performs relatively well with any pre-trained model.  %needs to attack process for each model, which is infeasible in practice. 
%Whereas, anonymized skeleton data with our proposed method has good performance to any pre-trained model.

\begin{table}[h!]
\centering
\begin{tabular}{cc|cc}
\toprule
\multicolumn{2}{c|}{Method}                                          & Action. & Iden.  \\ \midrule
\multicolumn{2}{c|}{Not-anonymized}                                 & 0.9510  & 0.8095 \\ \midrule
\multirow{6}{*}{Random noise}                 & $\sigma$ = 0.001 & 0.7565  & 0.7450 \\
                                    & $\sigma$ = 0.005 & 0.4430  & 0.3240 \\
                                    & $\sigma$ = 0.010 & 0.2660  & 0.1735 \\
                                    & $\sigma$ = 0.020 & 0.1265  & 0.1020 \\
                                    & $\sigma$ = 0.050 & 0.0455  & 0.0840 \\
                                    & $\sigma$ = 0.100 & 0.0450  & 0.0715 \\ \midrule
\multirow{2}{*}{Adversarial attack} & Attacked                      & 0.9435  & 0.0000 \\
                                    & Non-Attacked                  & 0.9435  & 0.3621 \\ \midrule
\multicolumn{2}{c|}{Our method}                                      & 0.9175     & 0.0420    \\ \bottomrule 
\end{tabular}
\caption{Comparison between our method and other alternative approaches. This experiment is conducted with the residual anonymizer on NTU60. }
\label{tbl:compare}

\end{table}

\begin{figure}[h]
     \centering
     \includegraphics[width=\columnwidth]{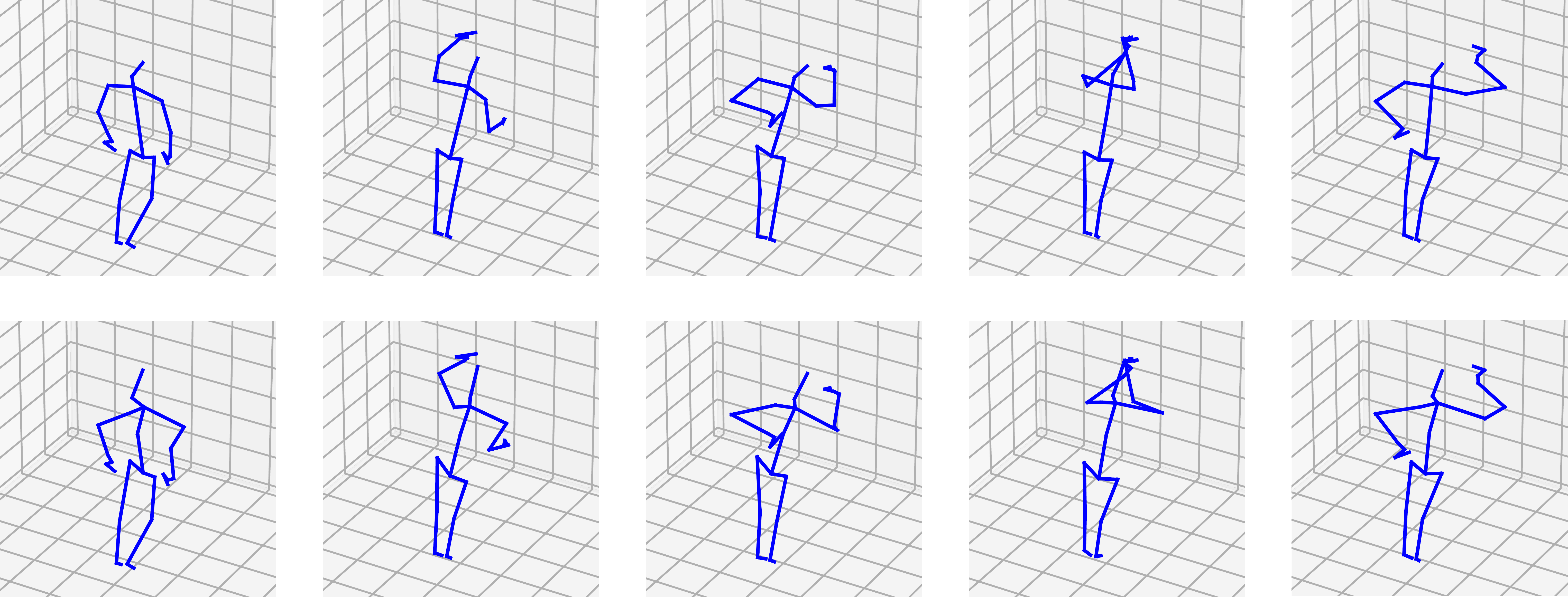}
     \caption{Five frames of the original (top) and the gender anonymized (bottom) skeletons for an action ``wiping face with a towel" from ETRI-activity3D. The subject is an elderly female.}
     \label{fig:qualitative_example}
 \end{figure}

\paragraph{Qualitative Analysis.}
To qualitatively understand the effect of anonymization, we visualize one example from the ETRI-activity3D dataset before and after anonymization in Figure \ref{fig:qualitative_example}. The top and bottom rows show five selected frames before and after anonymization for each figure, respectively. We can find some interesting patterns in the visualization. For example, the length of the neck bone is slightly increased, and the bone is moved to the upright position after anonymization. Given that an elderly female acts, we can conjecture that the adjustment makes gender unrecognizable. More visualization results are provided in Appendix.

\section{Related work}
Our work lies in public data anonymization and skeleton-based action recognition. In this section, we provide the previous attempts at dataset anonymization and skeleton-based action recognition. Also, we mention other research about adversarial attacks in skeleton data

\subsection{Public dataset anonymization}
%\paragraph{Public dataset anonymization.}
Researchers have pointed out privacy issues with public visual datasets and tried to mitigate them. \citet{caesar2020nuscenes, frome2009large, yang2021study} propose a blurring approach where the privacy-sensitive regions are blurred with an object detection method. \citet{flores2010removing, uittenbogaard2019privacy} propose an inpainting method to remove potentially problematic objects such as pedestrians and vehicles. A large body of prior work has used GANs\citep{goodfellow2014generative} to preserve visual private information. \citet{ren2018learning, maximov2020ciagan, hukkelaas2019deepprivacy} use GANs to generate fake faces to replace real ones. Also, \citet{gu2020password} proposes a face identity transformer that anonymizes face information according to the given password.

There are also some works that exist for other domains: sound domain~\citep{cohen2019voice, sumer2020automated} and text domain~\citep{li2018towards, coavoux2018privacy, mosallanezhad2019deep}. Similar concerns are also made for skeleton datasets. \citet{sinhaperson} propose a method to recognize persons from skeleton data. This work focuses on gait patterns extracted from human skeletons. The authors build a model with some predefined features and tested an adaptive neural network and na\"ive Bayes classifier to recognize the identity of persons. This implies the potential privacy leakage from public datasets.

\subsection{Skeleton-based action recognition}
Human skeleton data is a sequence of graphs, where joints and bones are represented as nodes and edges separately within a graph. In early times, skeleton motion trajectories are embedded into a manifold space as points. The relative distances between these points acted as clues for action recognition. However, these models do not exploit the internal spatial relationship between joints. Later, convolution neural networks (CNNs) are utilized to extract spatial co-occurrence patterns between joints. Nevertheless, CNNs cannot model a skeleton's topological information. 

Then, graph convolution networks (GCNs) are introduced to model these topological relations. Nonetheless, basic GCNs are not suitable for human skeleton sequences because they contain not only the 3D position of joints but also the time series. Yan et al. introduce the spatial-temporal graph convolutional networks (ST-GCN)~\cite{yan2018spatial}. They conduct graph convolution for extracting spatial features and perform  $1 \times 1$ convolution over each joint for capturing temporal variations. Following this line, various graph neural architectures are proposed to extract features from the graphs. 
AS-GCN~\cite{2019_cvpr_as_gcn} applies parametric adjacency matrices to substitute for the fixed skeleton graph. 
Subsequently, AGC-LSTM~\cite{2019_cvpr_agc_lstm} incorporates graph convolution layers into a long short-term memory network (LSTM) as gate operations to capture long-range temporal movements in action sequences.
The 2s-AGCN model~\cite{2019_cvpr_2sagcn} proposes bone features and learnable residual masks to enhance more flexibly extracting skeletons' structural information and ensembles the models trained separately with joints and bones to improve the classification accuracy. 
MS-G3D~\cite{Liu_2020_CVPR} introduces cross-spacetime skip connections, which additionally connect all 1-hop neighbors across all time frames of a dilated sliding window for direct information flow, and show improved recognition performance. Shift-GCN~\cite{cheng2020skeleton} applies shift graph operations and lightweight point-wise convolutions to overcome computational complexity and inflexible receptive fields of prior GCN-based studies.

In this work, we use Shift-GCN\citep{cheng2020skeleton}, MS-G3D\citep{Liu_2020_CVPR}, and 2s-AGCN\citep{2019_cvpr_2sagcn} as a baseline recognition model for private information. Although the original model is developed to recognize the actions of skeletons, we empirically show the model can successfully classify private information with a proper training procedure.

\subsection{Adversarial attacks in skeleton data}
Although deep neural network models achieve high performance in many tasks, there are main concerns about robustness. Researchers have investigated that imperceptible perturbed input data can easily deceive the deep neural network\citep{athalye2018synthesizing}. Recently, adversarial attacks have been attempted at skeleton-based action recognition\citep{liu2020adversarial, wang2021understanding, diao2021basar, tanaka2022adversarial, zheng2020towards}. These researches show that adversarial attacks successfully fool the state-of-art models with small perturbations. In this work, we use the adversarial attack method to attack privacy information and compare the result with our method.

%\vspace{-1mm}
\section{Conclusion}
\label{sec:conclusion}
% In this work, we investigate the privacy leakage from publicly available skeleton datasets.
% We propose a learning framework based on a neural architecture that reduces private information while preserving its usefulness. 
% % We propose a learning framework. It is a neural architecture that removes information about privacy while preserves movement patterns. 
% % based on a neural architecture that reduces private information while preserving its usefulness. 
% %Two learning algorithms are proposed, and their performances are measured on gender classification and re-identification tasks.
% As far as we know, our framework is the first gradient-based skeleton anonymization approach. 
% % As far as we know, the proposed framework is the first gradient-based skeleton anonymization approach. 
% We hope many novel and practical approaches can be made. 
% % We believe the proposed method opens new avenues to numerous applications. 
% For example, one can use rotation and translation invariant neural network architecture for the anonymizer such as the ones in~\cite{zhang2019rotation,pmlr-v139-satorras21a}. We leave this for future work.
%\vspace{-1mm}
In this work, we investigate privacy leakage from publicly available skeleton datasets. We show that although skeleton data may seemingly be privacy-protective, recently proposed skeletal action recognizers are surprisingly capable of extracting sensitive and identity information from these data. To address this privacy leakage problem, we propose a learning framework. Our experimental results reveal that the proposed method effectively removes the privacy information while preserving the movement patterns. 
Note that the anonymizers used in this work employ relatively simple architectures. Experiments show that private information can be removed effectively even with simple architectures.  
We leave the study of more advanced architectures for future work since our goal is to show the potential vulnerability of the skeletons and to provide a general framework to overcome.
%\sm{add limitation of our work}

\section*{Acknowledgement}
This work was partly supported by Institute of Information \& communications Technology Planning \& Evaluation (IITP) grant funded by the Korea government (MSIT) (No.2019-0-01906, Artificial Intelligence Graduate School Program(POSTECH)) and National Research Foundation of Korea (NRF) grant funded by the Korea government (MSIT) (NRF-2021R1C1C1011375). Dongwoo Kim is the corresponding author.

\bibliography{aaai23.bib}
% alpha and beta here is not the same to the code.
% in the code alpha, beta, and gamma were used,
% for the consistency,
% lr max-step is multiplied by alpha,
% and beta and gamma are divided by alpha (and renamed to alpha and beta respectively)

%\onecolumn

\clearpage

\section{Privacy Leakage}
\label{sec:privacy_leakage_result}

Table \ref{tbl:privacy_leakage} provides the detailed results of privacy leakage experiments. We use three baseline models: Shift-GCN\footnote{https://github.com/kchengiva/Shift-GCN}, MS-G3D\footnote{https://github.com/kenziyuliu/MS-G3D}, and 2s-AGCN\footnote{https://github.com/lshiwjx/2s-AGCN}. In this experiment, we set all hyperparameter configurations as default values given baseline models. To obtain a stable result, multiple training with random initialization is conducted with two NVIDIA GeForce RTX 3090 or two NVIDIA RTX A5000.

\begin{table}[h!]
\centering
\begin{tabular}{c|cc|c}
\toprule
\multirow{2}{*}{} & \multicolumn{2}{c|}{Re-identification} & Gender \\
                  & Top-1                & Top-5                & Accuracy              \\ \midrule
Shift-GCN         & $79.62_{\pm.70}$&$96.81_{\pm.09}$& $85.99_{\pm.40}$        \\
MS-G3D            & $82.23_{\pm.87}$ &$97.51_{\pm.07}$ & $87.90_{\pm.17}$  \\
2s-AGCN           & $76.89_{\pm1.83}$ & $96.56_{\pm.32}$& $86.43_{\pm.47}$  \\
\bottomrule        
\end{tabular}
\caption{Top-1 and Top-5 accuracy of re-identification task with NTU60 dataset, and accuracy of gender classification with ETRI-activity3D dataset. The re-identification task achieves 80\% and 97\% for top-1 and top-5 accuracy on average. Also, the gender classifier achieves 87\% accuracy on average.}
\label{tbl:privacy_leakage}
\end{table}

\section{Detailed Experimental Results}
\label{sec:experiment}
In this section, detailed hyperparameter configurations and results of our main experiments are provided. The code used in our experiments is available at \url{https://github.com/ml-postech/Skeleton-anonymization/}. Since the implementation is similar for all baselines, we only provide the code based on Shift-GCN. Also, data related to ETRI-activity3D will not be provided, since ETRI-activity3D can access by authorized users. If one wants to use this dataset for research purposes, please visit here\footnote{https://ai4robot.github.io/etri-activity3d-en/} and get permission. Additionally, we use U-Net architecture from Pytorch-UNet repository\footnote{https://github.com/milesial/Pytorch-UNet} for our anonymizer network. For all experiments, we set $k=1$ (see in Algorithm \ref{alg:adv}), i.e., one minimization with one maximization. 
We use four NVIDIA GeForce RTX 3090 or four NVIDIA RTX A5000 for training.

\paragraph{Anonymization results.}
Table \ref{tbl:final_identification} and Table \ref{tbl:final_gender} provide the detailed hyperparameter configurations and the entire results of Figure \ref{fig:final_results}. For all experiments, we train Shift-GCN to find a representative model at first. After finding the best $\alpha$ and $\beta$, the same $\alpha$ and $\beta$ are applied to the other two baselines with different learning rates. As the result, we set $\alpha$ and $\beta$ as 1.0 and 10 at ResNet, 0.3 and 2 at U-Net for NTU60. Also, we set $\alpha$ and $\beta$ as 2.0 and 15 at ResNet, 0.3 and 0.5 at U-Net for ETRI-activity3D, respectively.

\begin{table}[h!]
\centering
\begin{tabular}{ccccc}
\toprule
                      Model      & $lr$ & Re-iden. & Action. & Recon. error   \\ \midrule
\multicolumn{1}{r|}{Shift-Res} & 0.0100        & 0.04202   & 0.9175 & 0.00860 \\
\multicolumn{1}{r}{Shift-U}  & 0.0100        & 0.05701   & 0.9145 & 0.01655 \\ \midrule
\multicolumn{1}{r|}{MS-Res} & 0.0005        & 0.07770   & 0.9215 & 0.01092 \\
\multicolumn{1}{r|}{MS-U}  & 0.0005        & 0.10110   & 0.9243 & 0.01115 \\ \midrule
\multicolumn{1}{r|}{2s-Res} & 0.0050        & 0.05889   & 0.8994 & 0.01446 \\
\multicolumn{1}{r|}{2s-U}  & 0.0050        & 0.03514   & 0.8908 & 0.07536 \\
\bottomrule
\end{tabular}
\caption{Detail hyperparameter configurations and results of our representative model with three baselines for anonymizing identity information using NTU60.}
\label{tbl:final_identification}
\end{table}

\begin{table}[h!]
\centering
\begin{tabular}{ccccc}
\toprule
                      Model      & $lr$ & Gender. & Action. & Recon. error   \\ \midrule
\multicolumn{1}{r|}{Shift-Res} & 0.010        & 0.5875   & 0.8863 & 0.00127 \\
\multicolumn{1}{r}{Shift-U}  & 0.010       & 0.5671   & 0.8708 & 0.01086 \\ \midrule
\multicolumn{1}{r|}{MS-Res} & 0.001        & 0.6426   & 0.9073 & 0.00108 \\
\multicolumn{1}{r|}{MS-U}  & 0.001        & 0.6046   & 0.9033 & 0.00403 \\ \midrule
\multicolumn{1}{r|}{2s-Res} & 0.005        & 0.4221   & 0.8758 & 0.00397 \\
\multicolumn{1}{r|}{2s-U}  & 0.005        & 0.5645   & 0.8563 & 0.00920 \\
\bottomrule
\end{tabular}
\caption{Detail hyperparameter configurations and results of our representative model with three baselines for anonymizing gender information using ETRI-activity3D.}
\label{tbl:final_gender}
\end{table}

\paragraph{Trade-off analysis results.}
Table \ref{tbl:grid_iden} and Table \ref{tbl:grid_gender} show the detailed results of Figure \ref{fig:ntu_grid_result}. We vary hyperparameter $\alpha$ and $\beta$ to observe the trade-off between action accuracy and privacy accuracy. For these experiments, we use Shift-GCN and set the learning rate as 0.01.

\paragraph{Reconstruction error analysis results.}
Table \ref{tbl:reconstruction} provides the detailed results of Figure \ref{fig:recon_analysis}. To obtain different levels of reconstruction error, we vary $\beta$ with fixed $\alpha$ as 1. For these experiments, we use Shift-GCN with NTU60 and set the learning rate as 0.01. Multiple training with random initialization is conducted for stable results.

\begin{table}[h!]
\centering
\begin{tabular}{c|r|rr|rr}
\toprule
\multicolumn{1}{c}{\multirow{2}{*}{$\beta$}} & \multicolumn{1}{c}{\multirow{2}{*}{Recon. error}} & \multicolumn{2}{c}{Action}                             & \multicolumn{2}{c}{Re-iden.}                  \\
\multicolumn{1}{c}{}                      & \multicolumn{1}{c}{}                              & \multicolumn{1}{c}{Acc.} & \multicolumn{1}{c}{$\sigma^2$} & \multicolumn{1}{c}{Acc.} & \multicolumn{1}{c}{$\sigma^2$} \\
\midrule
5                                         & 0.012920                                           & 0.9144                       & 0.00203                & 0.08016                      & 0.01612                 \\
10                                        & 0.008804                                          & 0.9138                       & 0.00515                & 0.03964                      & 0.00680                \\
25                                        & 0.004570                                           & 0.9210                        & 0.00478                 & 0.08288                      & 0.03034                 \\
50                                        & 0.002268                                          & 0.9327                       & 0.00241                & 0.12080                       & 0.03837                 \\
75                                        & 0.001636                                          & 0.9385                       & 0.00255                & 0.15790                       & 0.04433        
 \\
\bottomrule
\end{tabular}
\caption{Detail results of reconstruction error analysis.}
\label{tbl:reconstruction}
\end{table}

\begin{table*}[hbp!]

\centering
\begin{tabular}{rrrrrr}
\toprule
$\alpha$  & $\beta$   & Action acc. & Re-iden. acc.  & Recon. error (RMSE)  &  Anonymization Network  \\
\midrule
1 & 5  & 0.9191 & 0.08303 & 0.00980 & ResNet\\
1 & 10 & 0.9175 & 0.04202 & 0.00860 & ResNet\\
1 & 15 & 0.9351 & 0.17070  & 0.00380 & ResNet\\
1 & 20 & 0.9235 & 0.12010  & 0.00550 & ResNet\\
2 & 5  & 0.8909 & 0.13190  & 0.02431 & ResNet\\
2 & 10 & 0.8838 & 0.09381 & 0.01822 & ResNet\\
2 & 15 & 0.9113 & 0.14690  & 0.01008 & ResNet\\
2 & 20 & 0.8728 & 0.07691 & 0.01020 & ResNet\\
3 & 5  & 0.8731 & 0.08266 & 0.02557 & ResNet\\
3 & 10 & 0.8814 & 0.07500   & 0.01340 & ResNet\\
3 & 15 & 0.8891 & 0.09286 & 0.01649 & ResNet\\
3 & 20 & 0.8595 & 0.10550  & 0.01410 & ResNet\\
0.3 & 0.5 & 0.9188 & 0.08657 & 0.02463 & U-Net \\
0.3 & 1   & 0.9206 & 0.07622 & 0.01945 & U-Net \\
0.3 & 1.5 & 0.9209 & 0.07891 & 0.01513 & U-Net \\
0.3 & 2   & 0.9145 & 0.05701 & 0.01655  & U-Net \\
0.5 & 0.5 & 0.9060  & 0.08045 & 0.02959 & U-Net \\
0.5 & 1   & 0.9074 & 0.06956 & 0.06496 & U-Net \\
0.5 & 1.5 & 0.9099 & 0.07791 & 0.02055 & U-Net \\
0.5 & 2   & 0.9071 & 0.05509 & 0.01750  & U-Net \\
0.7 & 0.5 & 0.8945 & 0.06597 & 0.12950  & U-Net \\
0.7 & 1   & 0.9022 & 0.05277 & 0.07635 & U-Net \\
0.7 & 1.5 & 0.8988 & 0.08494 & 0.02410  & U-Net \\
0.7 & 2   & 0.9006 & 0.04352 & 0.02232 & U-Net \\
\bottomrule
\end{tabular}
\caption{Results with varying hyperparameters for analyzing the trade-off between action accuracy and re-identification accuracy with NTU60.}
\label{tbl:grid_iden}
\end{table*}

\begin{table*}[hbp!]
\centering
\begin{tabular}{rrrrrr}
\toprule
$\alpha$  & $\beta$   & Action acc. & Gender acc.  & Recon. error (RMSE)  &  Anonymization Network  \\
\midrule
1 & 5  & 0.9028 & 0.6648 & 0.0017160 & ResNet \\
1 & 10 & 0.9032 & 0.6778 & 0.0007986 & ResNet\\
1 & 15 & 0.9050  & 0.7643 & 0.0004499 & ResNet\\
1 & 20 & 0.9032 & 0.6655 & 0.0003264 & ResNet\\
2 & 5  & 0.8940  & 0.6569 & 0.0061400   & ResNet\\
2 & 10 & 0.8925 & 0.6296 & 0.0023480  & ResNet\\
2 & 15 & 0.8863 & 0.5875 & 0.0012730  & ResNet\\
2 & 20 & 0.9011 & 0.7352 & 0.0011160    & ResNet\\
3 & 5  & 0.8890  & 0.6612 & 0.0061500   & ResNet\\
3 & 10 & 0.8934 & 0.6512 & 0.0035020  & ResNet\\
3 & 15 & 0.9031 & 0.7601 & 0.0025420  & ResNet\\
3 & 20 & 0.8958 & 0.6558 & 0.0020570 & ResNet\\
0.3 & 0.5 & 0.8708 & 0.5671 & 0.0108600  & U-Net \\
0.3 & 1   & 0.8754 & 0.5798 & 0.0094480 & U-Net \\
0.3 & 1.5 & 0.8337 & 0.5892 & 0.0264000   & U-Net \\
0.3 & 2   & 0.8758 & 0.5952 & 0.0108300  & U-Net \\
0.5 & 0.5 & 0.8569 & 0.5797 & 0.0156400  & U-Net \\
0.5 & 1   & 0.8303 & 0.5905 & 0.0276500  & U-Net \\
0.5 & 1.5 & 0.8341 & 0.5853 & 0.0249800  & U-Net \\
0.5 & 2   & 0.8567 & 0.5823 & 0.0117300  & U-Net \\
0.7 & 0.5 & 0.8239 & 0.6053 & 0.0294500  & U-Net \\
0.7 & 1   & 0.8265 & 0.5978 & 0.0326100  & U-Net \\
0.7 & 1.5 & 0.8269 & 0.5984 & 0.0277300  & U-Net \\
0.7 & 2   & 0.7504 & 0.6046 & 0.0469700  & U-Net\\
\bottomrule
\end{tabular}
\caption{Results with varying hyperparameters for analyzing the trade-off between action accuracy and gender accuracy with ETRI-activity3D.}
\label{tbl:grid_gender}
\end{table*}

\section{Hyperparameter Analysis}
\begin{figure}[h!]
\centering
\begin{subfigure}[h]{0.45\textwidth}
\includegraphics[width=\linewidth]{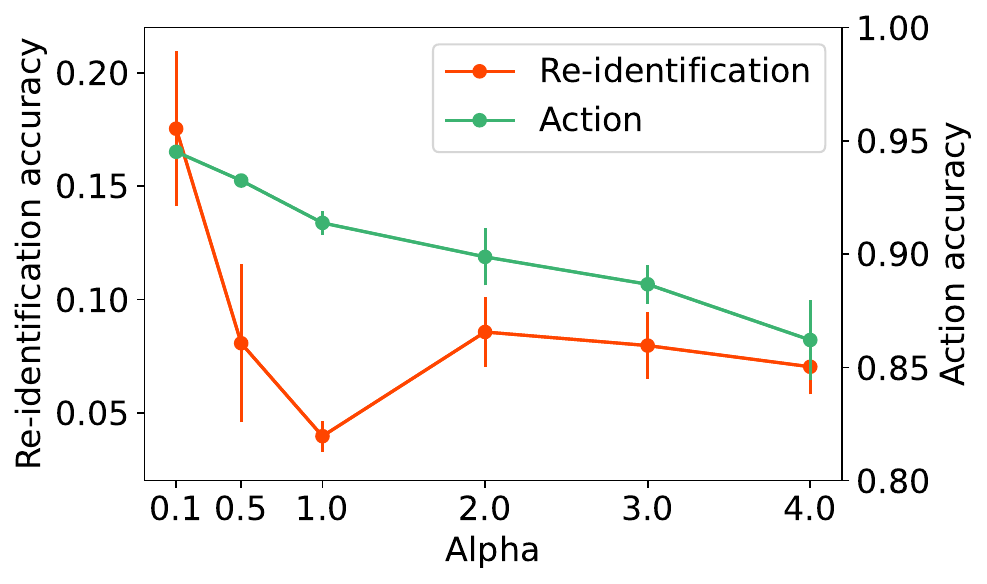}
\caption{\label{fig:alpha_analysis}Re-identification and Action accuracy}
\end{subfigure}
\begin{subfigure}[h]{0.4\textwidth}
\includegraphics[width=\linewidth]{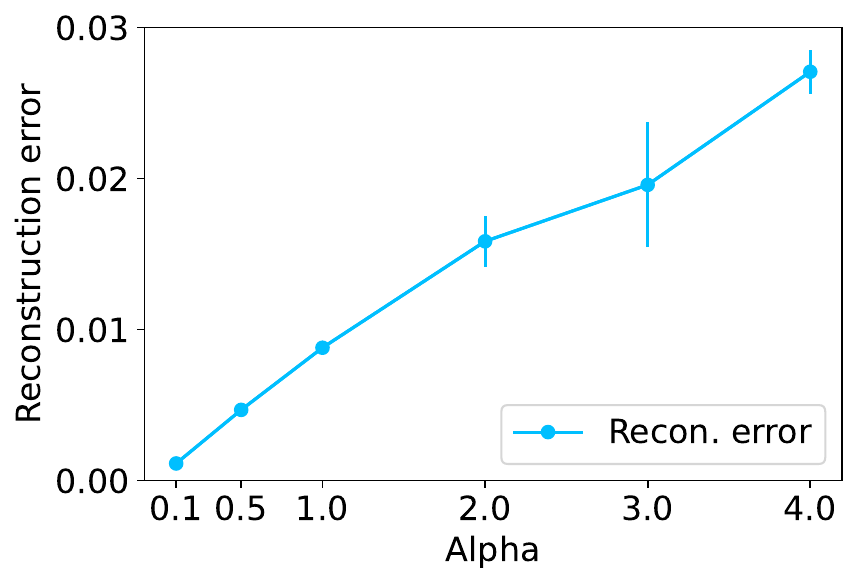}
\caption{\label{fig:alpha_analysis_recon}Reconstructsion error}
\end{subfigure}

\caption{Re-identification accuracy, action accuracy, and reconstruction error with different $\alpha$}
\label{fig:alpha_analysis_result}
\end{figure}

In this section, we analyze the hyperparameter used in our objective function (See Equation \ref{eqn:minimize}) with action accuracy, re-identification accuracy, and reconstruction error. There are two hyperparameters:  $\alpha$ and $\beta$. $\alpha$ adjusts the importance of privacy, one expects that large $\alpha$ may occur low privacy accuracy. Also, $\beta$ adjusts the reconstruction error, which regularizes anonymized data to look similar to original data. We conduct this experiment with Shift-GCN, NTU60, and ResNet.
\paragraph{Alpha term analysis.}
Figure \ref{fig:alpha_analysis_result} is the result of different $\alpha$. In this experiment, we vary $\alpha$ with fixed $\beta$ of 10. As shown in Figure \ref{fig:alpha_analysis}, action accuracy decreases, as $\alpha$ increases. Also, re-identification accuracy drops rapidly from 0.1 to 1.0, but increases at 2.0 and decreases again. Note that $\alpha$: 1, $\beta$: 10 is the best model in our metric. This means that optimizing the minimax objective works well under a certain $\alpha$ so that growing $\alpha$ produces low re-identification accuracy. However, $\alpha$ that exceeds a particular value disturbs optimization due to a trade-off with action accuracy and reconstruction error. Also, Figure \ref{fig:alpha_analysis_recon} shows that reconstruction error increases, as $\alpha$ increases. Since the ratio of $\beta$ to $\alpha$ decreases, so regularizing anonymized data can not influence enough during training.

\begin{figure}[h!]
\centering
\begin{subfigure}[h]{0.45\textwidth}
\includegraphics[width=\linewidth]{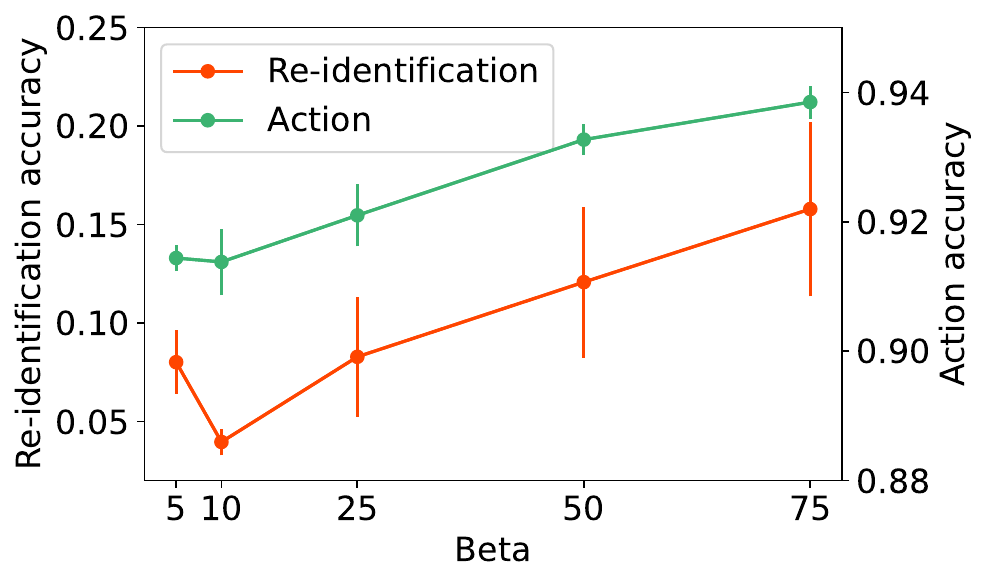}
\caption{\label{fig:beta_analysis}Re-identification and Action accuracy}
\end{subfigure}
\begin{subfigure}[h]{0.4\textwidth}
\includegraphics[width=\linewidth]{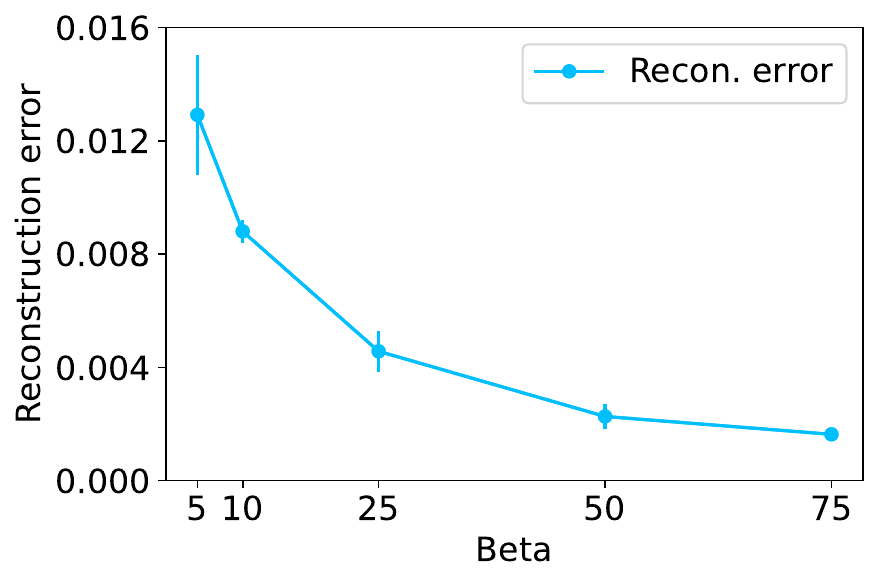}
\caption{\label{fig:beta_analysis_recon}Reconstruction error}
\end{subfigure}

\caption{Re-identification accuracy, action accuracy, and reconstruction error with different $\beta$}
\label{fig:beta_analysis_result}
\end{figure}

\paragraph{Beta term analysis.}
Figure \ref{fig:beta_analysis_result} is the result of according to different $\beta$. In this experiment, we vary $\beta$ with fixed $\alpha$ of 1. As shown in Figure \ref{fig:beta_analysis_recon}, it is trivial that reconstruction error increases by raising $\beta$. This means that large $\beta$ makes anonymized data look similar to original data. This affects re-identification accuracy and action accuracy. Figure \ref{fig:beta_analysis} shows that action accuracy and re-identification accuracy increase, as $\beta$ increases. 

\section{Qualitative Analysis}
\label{sec:Quantitative}
Figure \ref{fig:qualitative_example} is visualization results of before and after anonymization with NTU60 and ETRI-activity3D dataset. Our qualitative results show that anonymized skeleton data success to remove privacy information while keeping action information with some interesting patterns. e.g. length of the neck bone is increased, bone is moved to the upright position, etc.
\begin{figure*}[!htb]
\centering
     \begin{subfigure}[b!]{\textwidth}
         \centering
         \includegraphics[width=\linewidth]{figures/figure6/etri_20247.pdf}
         \caption{The original (top) and the gender anonymized (bottom) skeletons for an action ``wiping face with a towel" from ETRI-activity3D. The subject is an elderly female. }
         \label{fig:first_example}
         %\vspace{0.5em}
     \end{subfigure}
     \begin{subfigure}[b!]{\textwidth}
         \centering
         \includegraphics[width=\linewidth]{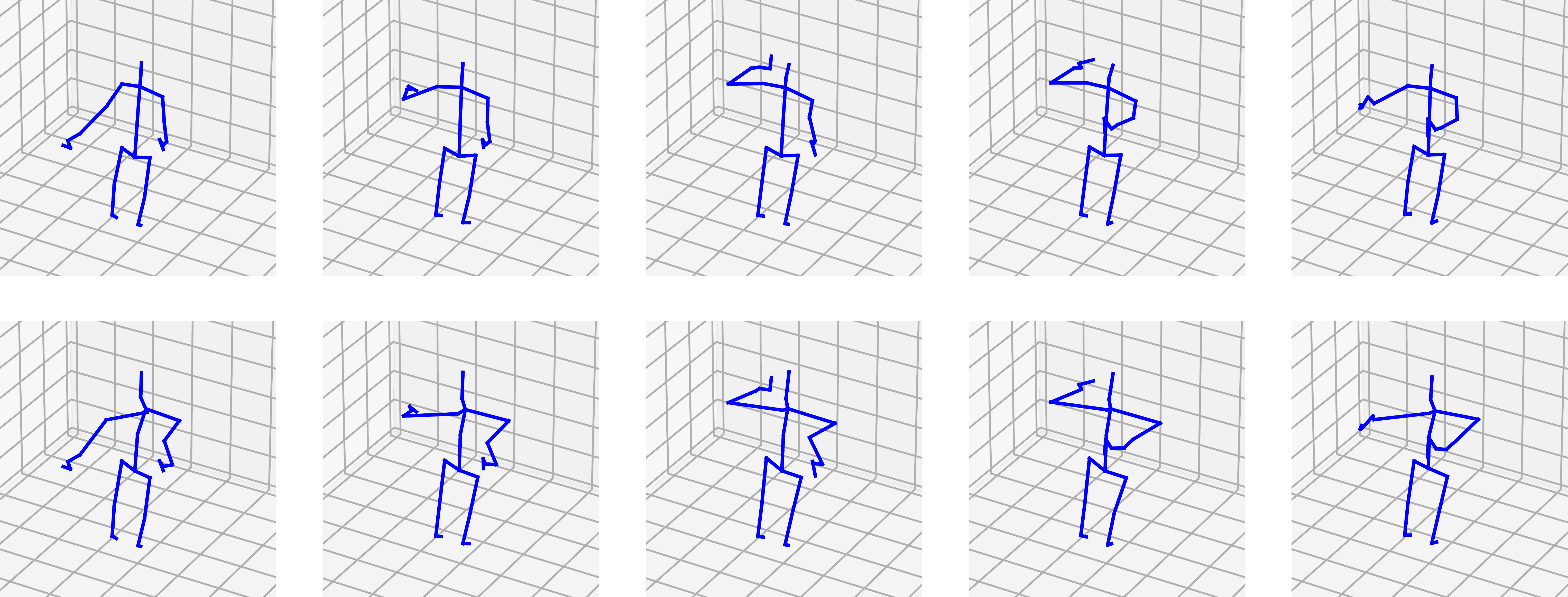}
         \caption{The original (top) and the gender anonymized (bottom) skeletons for an action ``drinking water" from ETRI-activity3D. The subject is an elderly male.}
         \label{fig:second_example}
         %\vspace{0.5em}
     \end{subfigure}
     \begin{subfigure}[b!]{\textwidth}
         \centering
         \includegraphics[width=\linewidth]{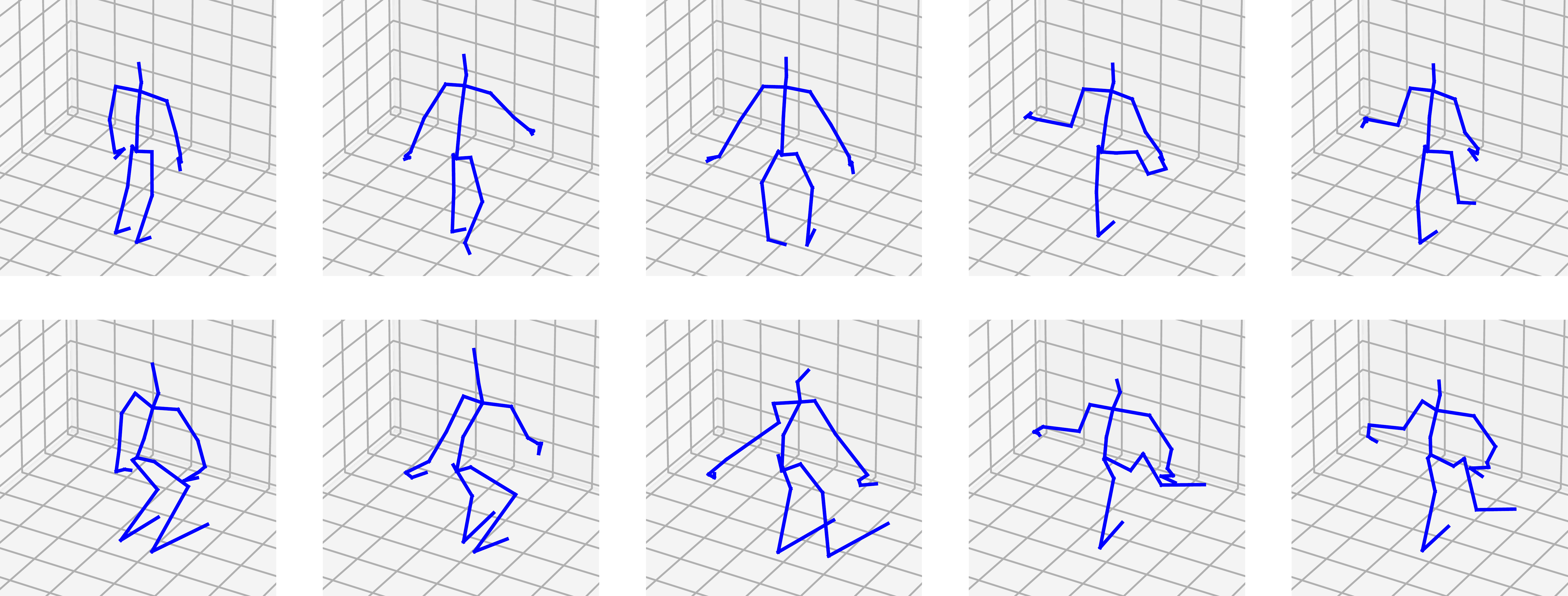}
         \caption{The original (top) and the identity anonymized (bottom) skeletons for an action ``kicking something" from NTU60. The subject's ID is 2.}
         \label{fig:third_example}
     \end{subfigure}
\caption{Examples of anonymized skeletons. Five frames are visualized from a sequence of action frames.} %Top: Before anonymization. Bottom: After anonymization.}
\label{fig:qualitative}
\end{figure*}

% \begin{table}[h!]
% \centering
% \begin{tabular}{cc|ccc}
% \toprule
% \multicolumn{2}{c|}{Model}          & Recon. & Action & Gender \\ 
% \midrule
% \multicolumn{2}{c|}{Not-anonymized} & 0.0000 & 0.9070 & 0.8890 \\ 
% \midrule
% \multirow{6}{*}{RN}     & $\sigma$ = 0.001     & 0.0010 & 0.7385 & 0.8645 \\
%                         & $\sigma$ = 0.005     & 0.0050 & 0.4000 & 0.7350 \\
%                         & $\sigma$ = 0.010     & 0.0100 & 0.1715 & 0.5810 \\
%                         & $\sigma$ = 0.020     & 0.0200 & 0.0710 & 0.4315 \\
%                         & $\sigma$ = 0.050     & 0.0500 & 0.0255 & 0.3860 \\
%                         & $\sigma$ = 0.100     & 0.1000 & 0.0180 & 0.4075 \\
% \bottomrule

% \end{tabular}
% \end{table}

\end{document}